\documentclass{article}
\usepackage{spconf,amsmath,graphicx,subfigure,epsfig}
\usepackage{multirow}
\usepackage{booktabs}

\def\x{{\mathbf x}}

\title{VISUAL AND TEXTUAL SENTIMENT ANALYSIS USING DEEP FUSION CONVOLUTIONAL NEURAL NETWORKS}
\name{Xingyue Chen \qquad Yunhong Wang \qquad Qingjie Liu\sthanks{Corresponding author} \qquad }
\address{
	The State Key Lab. of Virtual Reality Technology and Systems,
	Beihang University, Beijing 100191, China\\
	}
\begin{document}
%
\maketitle
\begin{abstract}
Sentiment analysis is attracting more and more attentions and has become a very hot research topic due to its potential applications in personalized recommendation, opinion mining, etc. Most of the existing methods are based on either textual or visual data and can not achieve satisfactory results, as it is very hard to extract sufficient information from only one single modality data. Inspired by the observation that there exists strong semantic correlation between visual and textual data in social medias, we propose an end-to-end deep fusion convolutional neural network to jointly learn textual and visual sentiment representations from training examples. The two modality information are fused together in a pooling layer and fed into fully-connected layers to predict the sentiment polarity. We evaluate the proposed approach on two widely used data sets. Results show that our method achieves promising result compared with the state-of-the-art methods which clearly demonstrate its competency.
\end{abstract}

\section{Introduction}
\label{sec:intro}
\vspace{-0.1cm}
Recently, with the rapid development of the Mobile Internet and smart terminals, more and more people prefer to use short texts and images to express their opinions and communicate with each other on social medias, such as Twitter, Flickr, Microblog, etc. Each day, billions of messages and posts are generated. Mining sentiment information from these data is of great important for applications such as personalized recommendation, user modeling, crisis management, etc~\cite{Wu2016}. It has been attracting more and more attentions and has become a hot research topic. Plenty of works have been published~\cite{Wu2016, pang2008opinion, machajdik2010affective, you2015robust, wang2016beyond, ren2016context}, significantly promoting the development of sentiment analysis studies.

Most of these methods are based on text information and consider sentiment analysis as a special case of text classification, focusing on designing specific vector representation of sentences~\cite{pang2008opinion}. For example, Le and Mikolov~\cite{le2014distributed} proposed an approach to learn distributed representations for documents and applied on sentiment analysis, achieving excellent performance. Recently, with the development of deep learning techniques, researchers tend to use deep neural networks to learn distributed and robust representations of text for sentiment classification.  Kim~\cite{kim2014convolutional} and Johnson~\cite{johnson2014effective} implemented Convolutional Neural Networks (CNN) for text classification. Poria et al.~\cite{poria2015deep} presented a novel way of extracting features from short texts, which is based on the activation values of an inner layer of a deep CNN. These deep learning based methods obtain better performance than hand-crafted features in text sentiment analysis.

Besides text data, visual content of images also convey sentiment information~\cite{wang2016beyond}. The main challenge behind visual sentiment analysis is how to learn effective representations to bridge the "affective gap" between low level features of images and high level semantics~\cite{machajdik2010affective}. Motivated by the fact that sentiment involves high-level abstraction, such as objects or attributes in images, Yuan et al.~\cite{yuan2013sentribute} proposed to employ visual entities or attributes as features for visual \mbox{sentiment} prediction. Borth et al.~\cite{borth2013large} employed 1,200 adjective noun pairs (ANP) which were used as keywords to crawl images from Flickr as mid-level features for sentiment classification. To handle large scale weakly labeled training images, You et al.~\cite{you2015robust} designed a progressively trained CNN architecture for visual sentiment analysis. Compared with approaches employing low-level or mid-level features, deep learning algorithms learn hierarchical features and can achieve outstanding performance.

\begin{figure}[t]\centering
	\subfigure[]
	{\centering
		\begin{minipage}[t]{0.22\textwidth}
			\includegraphics[width=4cm]{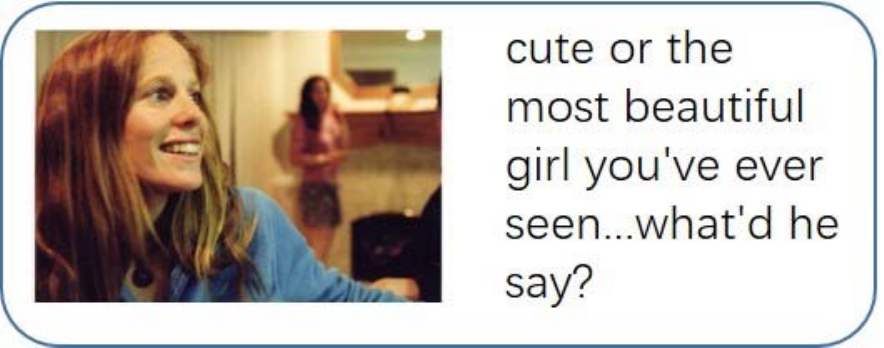}
            \vspace{-0.2cm}
			\label{fig1:a}
		\end{minipage}
	}
	\vspace{-0.1cm}
	\subfigure[]
	{\centering
		\begin{minipage}[t]{0.22\textwidth}
			\includegraphics[width=4cm]{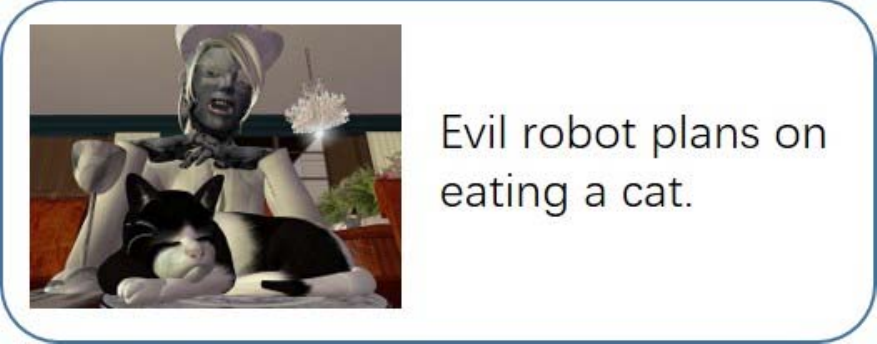}
            \vspace{-0.2cm}
			\label{fig1:b}
		\end{minipage}
	}
	\subfigure[]
	{\centering
		\begin{minipage}[t]{0.22\textwidth}
			\includegraphics[width=4cm]{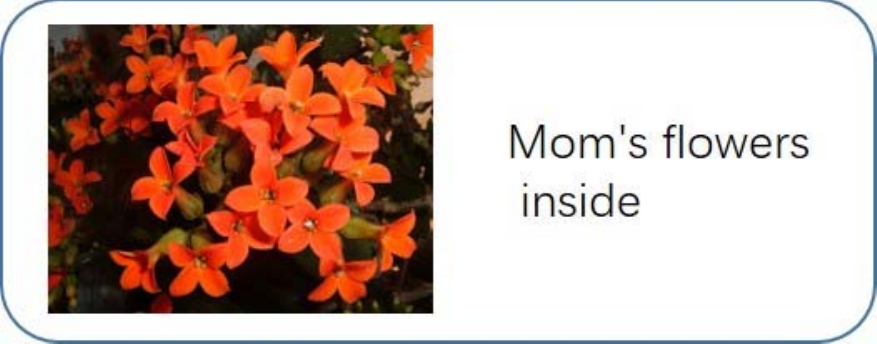}
            \vspace{-0.2cm}
			\label{fig1:c}
		\end{minipage}
	}
	\subfigure[]
	{\centering
		\begin{minipage}[t]{0.22\textwidth}
			\includegraphics[width=4cm]{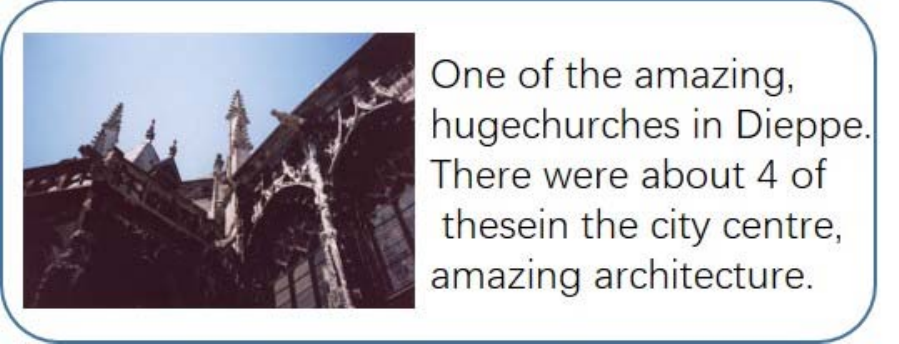}
            \vspace{-0.2cm}
			\label{fig1:d}
		\end{minipage}
	}
	\caption{Example image and co-related text pairs crawled from Flickr.}
	\label{fig:prossess4}
	\vspace{-0.3cm}
\end{figure}

In social medias, one message or post usually contains both images and texts. They tend to have a high degree of relevance in semantic space, providing complementary information for each other. So, it is better to take both of them into consideration when determining sentiment. Fig.~\ref{fig:prossess4} shows several examples of image and text pair crawled from Flickr\footnotemark[1]. It can be observed from these four examples that, both image and the corresponding text in Fig.~\ref{fig1:a} indicate that this tweet carries a positive sentiment; while in Fig.~\ref{fig1:b} both are negative; in Fig.~\ref{fig1:c}, it is difficult to tell the sentiment from the text, however, we can tell that this image expresses positive sentiment; on the contrary, in Fig.~\ref{fig1:d}, it is hard to tell the sentiment from the image, however the word 'amazing' in the text suggests an overall positive sentiment. These examples explain the motivation for our work. Since the sentiment conveyed from visual and textual content can explain or even strengthen each other, we learn people's sentiment through the available images and the informal short text.
\footnotetext[1]{https://www.flickr.com/}

\begin{figure*}[tbp]
	\centering
	\includegraphics[width=0.9\textwidth]{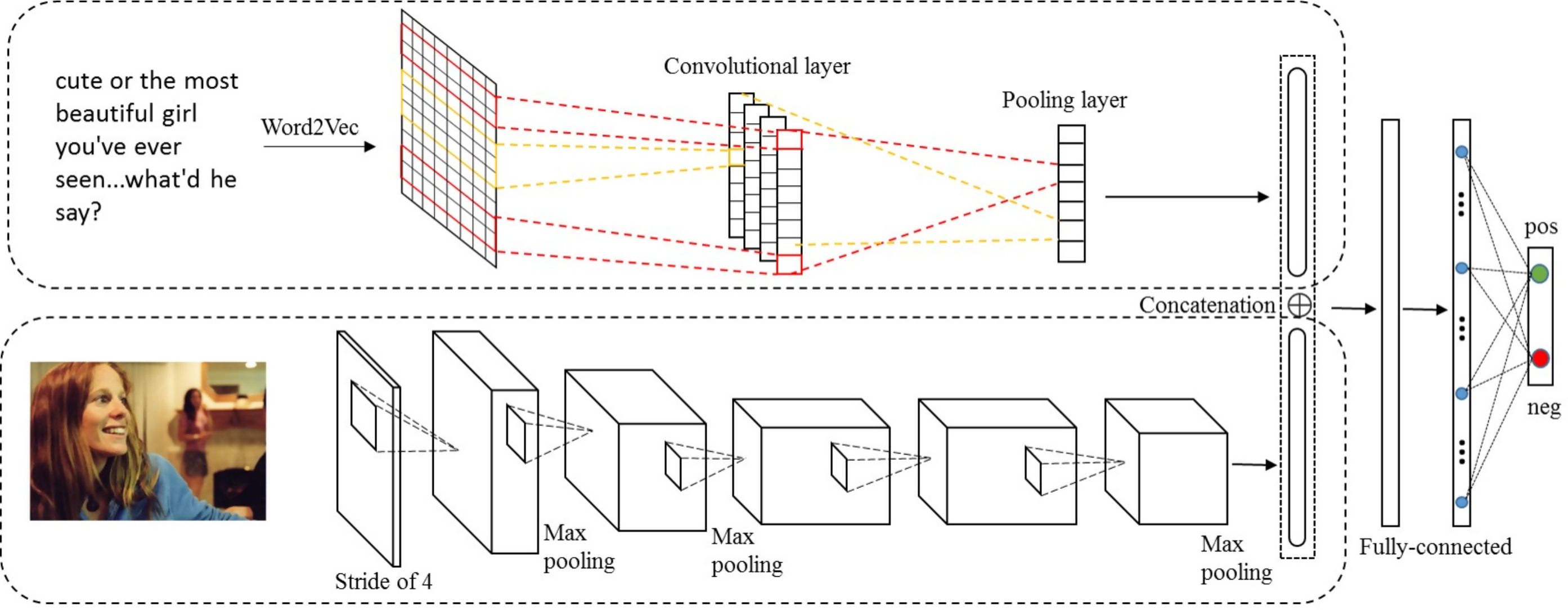}
	\caption{Framework of the proposed approach.}\label{fig:sys}
	\vspace{-0.3cm}
\end{figure*}

Recent progress in sentiment classification proves the advantage of using both text and visual modalities. Wang et al.~\cite{wang2014microblog} proposed a decision level fusion method for sentiment classification, which combines textual sentiment obtained by N-gram features with visual sentiment obtained with mid-level visual feature to predict final results. Later, You et al.~\cite{you2016cross} proposed a cross-modality consistent regression (\mbox{CCR}) scheme for joint textual-visual sentiment analysis. Their approach employed deep visual and textual features to learn a regression model. Despite effectiveness, these models ignore the intrinsic relationship between text and images and use the visual and textual information separately. To tackle this, You et al.~\cite{you2016robust} employed a tree-structured LSTM for textual-visual sentiment analysis, leading to better mapping between textual words and image regions. This model achieved the best performance over other fusion models. Although this model makes better use of the correspondence between text and images, it lost the global information of image in some extent, which we think is also important for visual sentiment.

Inspired by the recent discovery that contextual information plays an import role in sentence classification~\cite{ren2016context}, we propose to utilize a structured model similar to \cite{ren2016context} for visual-textual sentiment analysis. Different from \cite{ren2016context}, we use image as contextual information source. Visual and textual sentiment representations are learned through multi-layer CNNs and fused together in a pooling layer to predict sentiment polarity. We carry out experiments on two public data sets, including the Visual Sentiment Ontology (VSO) dataset~\cite{borth2013large} and the Multilingual Visual Sentiment Ontology (MVSO) dataset~\cite{jou2015visual}. Our method achieves the state of the art result on VSO dataset with 1.3\% accuracy gain, demonstrating its effectiveness.

The rest of this paper is organized as follows, Section 2 details the proposed approach. We conduct experiments on Section 3. And finally the paper is concluded in Secton 4.
\section{Deep fusion sentiment classification}
\label{sec:format}
In this section, we introduce the proposed deep fusion neural network for visual and textual sentiment analysis. The architecture of the proposed network is shown in Fig.~\ref{fig:sys}, which has two sub-networks. The upper sub-network takes text as input, and extracts vector representations of words. While the lower sub-network takes image as input and extracts hierarchical features of image. The two representations are combined together in the following pooling layer to make full use of two modalities information.

 \subsection{Deep visual sentiment feature}
 \label{ssec:subhead}
The recent developed CNN has been proved to be effective in vision tasks such as image classification and object detection. As visual sentiment analysis can be treated as a high-level image classification problem, You et al.~\cite{you2015robust} applied CNN to sentiment predication and achieved better performance than using traditional low-level and mid-level features. Inspired by the success of CNN and You's work, in this paper, we also implement a CNN architecture to learn high level visual sentiment representations.

Although, more layers in CNN generally indicate a better performance, it has been reported that going deeper may lead to worse result for visual sentiment analysis~\cite{you2015robust}. And shallow networks may not be deep enough to learn high level semantic representations. So, in our approach, we employ a network that is similar to AlexNet~\cite{krizhevsky2012imagenet}, both consisting of 5 convolutional layers and taking a 244$\times$244 pixels RGB image as input. The nonlinearity activation function of each neuron in this CNN is modeled by Rectified Linear Units (ReLUs),
   \begin{equation}
    f(x)=\max(0, x).
    \end{equation}
Compared with saturating nonlinearity function, such as sigmoid, ReLUs can accelerate learning process, and prevent the gradients vanishing during back propagation.

Each convolutional layer convolves the output of its previous layer with a set of learned kernels, followed by ReLU non-linearity, and two optional layers: local response normalization and max pooling. The local response normalization layer is applied across feature channels, and the max pooling layer is applied to neighboring neurons. CNN learns hierarchical feature representations. One simple way to use these features is making use all of them. However, since sentiment analysis involves high level abstraction, we take the last output of max pooling layer as the visual sentiment representation.
  \subsection{Learning textual sentiment representation}
 \label{ssec:subhead}

Recently, deep learning models are also successfully applied in many natural language processing tasks,  such as sentence classification~\cite{kim2014convolutional}. We use the same architecture as \cite{kim2014convolutional} to learn textual sentiment representation.

In our work, we directly employ a pre-trained distributed representations of words, keep the word vectors static and learn only the other parameters of the model.
Let $x_i \in R^k$ be a $k$-dimensional word vector corresponding to the $i$-th word in the sentence. A sentence of length $n$ is represented as
  \begin{equation}
  x_{1:n}= x_1 \oplus x_2 \oplus \cdots \oplus x_n
  \end{equation}
where $\oplus$ is the concatenation operator. For convenience, let $x_{i:i+j}$ represents the concatenation of words $x_i, x_{i+1},  . . . , x_{i+j}$. A convolution operation involves a filter $w \in R^{hk}$,  which is applied to a window of $h$ words to produce a new feature. For example, a feature $c_i$ is generated from a window of words $x_{i:i+h-1}$ by
  \begin{equation}\
    c_i = f(w * x_{i:i+h-1}+b)
  \end{equation}
Here * is the convolution operation, $b \in R$ is a bias term and $f$ is a non-linear function such as the hyperbolic tangent. This filter is applied to each possible window of words in the sentence $\left\{x_{1:h}, x_{2:h+1}, \cdots, \\x_{n-h+1:n}\right\}$ to produce a feature map
  \begin{equation}\
    \textbf{c} = [c_1, c_2, \cdots, c_{n-h+1}]
  \end{equation}
We exploit pooling techniques to merge the varying number of features from the convolution layer into a vector with fixed dimensions. A typical pooling technique is the max pooling, which chooses the highest value on each dimension from a set of vectors. On the other hand,  min and average pooling have also been used for sentiment classification\cite{vo2015target}, giving significant improvements. We consider all the three pooling methods,  concatenating them together as a new hidden layer $\hat{h}$. Formally, the values of $\hat{h}$ are defined as:
  \begin{equation}\
    \hat{h} = \left[ \max\left\{\textbf{c}\right\}, \mathrm{avg}\left\{\textbf{c}\right\}, \min\left\{\textbf{c}\right\} \right]
  \end{equation}
  For one filter , we get a three-dimensional vector and each dimension represents the output of one pooling method. We have described the process by which one feature is extracted from one filter. Our model uses multiple filters (with varying window sizes) to obtain features. The $\hat{h}$ of the different filters is concatenated in the pooling layer. We take these features as the textual sentiment representation.
  \subsection{Deep fusion for sentiment analysis}
 \label{ssec:subhead}
In social medias, text and image often appear in pairs, providing complementary semantic information for each other. Ren et al.~\cite{ren2016context} have proved that context information can be helpful in social text sentiment analysis. Inspired by their work, we  believe the co-related image can also provide contextual information and is beneficial for improving sentiment analysis performance. To this end, we build a fusion layer,  combining the output of image pooling layer with that of the text neural model, before feeding them to the non-linear hidden layer. As the output of textual pooling layer is high dimensional feature vector, the output of visual pooling layer should be reshaped to one-dimensional vector. The fused feature is represented as
  \begin{equation}
  x = x_i \oplus x_t
  \end{equation}
where $\oplus$ is the concatenation operator. $x_i$ and $x_t$ refer to the output of two sub-networks, respectively. In this way, the hidden layer automatically combines visual and textual sentimental representations.

The fusion layer is followed by three fully-connected layers. The output of the last fully-connected layer is fed to a 2-way softmax which produces a distribution over the 2 class labels. Our training objective is to minimize the cross-entropy loss over a set of training examples:
  \begin{equation}
  L_i = -\log \left( \frac{e^{f_{y_i}}}{\sum_je^{f_j}} \right)
  \end{equation}
where we use the notation $f_j$ to mean the $j$-th element of the vector of class scores $f$.

\section{Experiments}
\label{sec:majhead}
We evaluate the proposed deep fusion neural network on two datasets and compare the performance of our method with 3 state of the art methods.
\vspace{-0.3cm}
  \subsection{Datasets}
 \label{ssec:subhead}
There are a few public datasets available for testing. And most of them are either texts or images. We use the following two largest datasets which can provide texts and images for evaluation.

\textbf{VSO}~\cite{borth2013large}. VSO is built on top of the visual sentiment ontology,  and contains millions of images collected by querying Flickr with thousands of adjective and noun pairs (ANPs). Each ANP has hundreds of images and the sentiment label of each image is decided by sentiment polarity of the corresponding ANP. The dataset provides the metadata which enables us to obtain the description of an image. We only use image with descriptions more than 5 words and less than 150 words, that leaves us 346,139 images. The training/testing split is 80\% and 20\%.

\textbf{MVSO-EN}~\cite{jou2015visual}. The MVSO dataset consists of 15,600 concepts in 12 \mbox{different} languages that are strongly related to emotions and sentiments expressed in images. Similar to the VSO dataset, these concepts are defined in the form of ANPs,  which are crawled from Flickr. In this paper, we only use the English dataset for our experiments. In order to avoid the problem of excessive noise,  we choose ANPs with sentiment scores more than 0.16 and less than -0.1. We skip images with descriptions more than 150 words and less than 5 \mbox{words}. There are total 613,955 image and text pairs used in our experiment.The training/testing split is 80\% and 20\%.




\vspace{-0.3cm}
  \subsection{Experimental results}
 \label{ssec:subhead}
To demonstrate the effectiveness of the proposed method, we first compare it with two baseline methods, that is only \mbox{using} image or text information for sentiment classification. We call them Single Visual (SV) and Single Text (ST) models. To show the superiority of our method, we further compare it with 3 state of the art methods, including PCNN~\cite{you2015robust}, You's method~\cite{you2016cross} and T-LSTM~\cite{you2016robust}. Specially, You et al. utilized three strategies, named early fusion, which is logistic regression on concatenated visual and textual features; late fusion, which is average of logistic regression sentiment {} score on both visual and textual features; and cross-modality consistent regression (CCR).


In the textual part of our method, we employ the pre-trained Word2Vec~\cite{mikolov2013distributed} model to get the distributed representations of words. The Word2Vec was pre-trained on Wikipedia Corpus, and has a fixed size of 200 dimension. Words do not present in pre-training dataset are initialized randomly. The size of embedded word matrix is 150$\times$200, as the max-length of description is 150 words. The first layer performs convolution over the embedded word vectors using scalable filter size. In this work,  we slide over 3, 4 or 5 words at one time.
In the visual part,  the input images are first resized to 224$\times$224 before inputing to the network. The first five layers are same as AlexNet~\cite{krizhevsky2012imagenet}.

The model was trained on a workstation with i7-5600 CPU, 32G RAM, NVDIA 1080 GPU. We trained our model using a mini-batch of 100. The initial learning rate is 0.0001 and exponential decays  every 3000 steps with a base of 0.96.

Table~\ref{tab:VSO} summarizes the performances of our approach and comparison methods. It can be observed that the \mbox{proposed} approach significantly outperforms baselines and PCNN method, indicating fusing two modalities information does improve sentiment prediction. This is mainly because multi-modalities information enrich the semantic space. Our deep fusion model also obtains better results than T-LSTM and \mbox{You's} method. We believe it mainly because our model utilizes the global information of image.
 \vspace{-0.4cm}
\begin{table}[h]
	\centering
	\small
	\caption{Results on VSO dataset.}
	\label{tab:VSO}
	\begin{tabular}{c|c|c|c|c|c}
        \hline
        \multicolumn{2}{c|}{\textbf{Methods}} & \textbf{Prec.} & \textbf{Rec.} & \textbf{F1} & \textbf{Acc.} \\
        \hline
        \multirow{2}{*}{Baseline}& ST& 0.701 & 0.680 & 0.690 & 0.695  \\
        \cline{2-6}
                                 & SV & 0.628 & 0.640 & 0.629 & 0.631   \\
        \hline
       \multicolumn{2}{c|}{PCNN~\cite{you2015robust}} & 0.759 & 0.826 & 0.791 & 0.781  \\\hline
        \multirow{3}{*}{You's~\cite{you2016cross}}&Early Fusion & 0.616 & 0.646 & 0.631 & 0.621  \\
        \cline{2-6}
        &Late Fusion& 0.660 & 0.629 & 0.645 & 0.650  \\
        \cline{2-6}
        &CCR & 0.672 & 0.678 & 0.675 & 0.672  \\\hline
        \multicolumn{2}{c|}{T-LSTM~\cite{you2016robust}} & 0.821 & 0.833 & 0.833 & 0.833  \\\hline
        \multicolumn{2}{c|}{ Deep fusion} & \textbf{0.830} & \textbf{0.857} & \textbf{0.844} & \textbf{0.847}  \\
        \hline
	\end{tabular}
	\vspace{-0.3cm}
\end{table}


Table~\ref{tab:MSVO} shows the performance of our deep fusion neural network on MVSO-EN dataset. To our best knowledge, this is the first performance report on MVSO data with all textual and visual information used. We used the same experimental setting to VSO, however obtained worse results than VSO. This is mainly because the MVSO-EN dataset contains much more noise than VSO.
 \vspace{-0.4cm}
\begin{table}[h]
	\centering
	\caption{Results of our method on MVSO-EN dataset.}
	\label{tab:MSVO}
	\begin{tabular}{c|c|c|c|c|c}
        \hline
        \multicolumn{2}{c|}{\textbf{Methods}} & \textbf{Prec.} & \textbf{Rec.} & \textbf{F1} & \textbf{Acc.} \\
        \hline
        \multirow{2}{*}{Baseline}& ST& 0.654 & 0.671 & 0.662 &  0.667 \\
        \cline{2-6}
                                 & SV & 0.602 &0.578  & 0.590 &  0.581 \\
        \hline
        \multicolumn{2}{c|}{ Deep fusion} & \textbf{0.740} & \textbf{0.730} & \textbf{0.735} & \textbf{0.737}  \\
        \hline

	\end{tabular}
	\vspace{-0.3cm}
\end{table}
\vspace{-0.3cm}
\section{Conclusions}
\vspace{-0.3cm}
\label{sec:ref}
In this paper, we present a new end-to-end framework for visual and textual sentiment analysis. Our method can effectively learn sentiment representations from noisy web images with text,  integrate both of them in a deep fusion layer and predicts an overall sentiment.
Extensive experimental results demonstrate that the proposed deep fusion convolutional neural network have significantly improved the performance of visual and textual sentiment analysis on VSO datasets, which shows its effectiveness for this issue.
\vspace{-0.4cm}
\section*{Acknowledgment}
\vspace{-0.3cm}
This work is surpported by the National Nature Science Foundation of China (61573045,61573045).
\bibliographystyle{IEEEbib}
\bibliography{strings,myrefs}

\end{document}